\documentclass{article}

\RequirePackage{hyperref}
\usepackage{graphicx}
\usepackage{paralist}
\usepackage{amsmath}

\begin{document}

\title{A Tutorial on Modular Ontology Modeling with Ontology Design Patterns:\\ The Cooking Recipes Ontology}
\author{Pascal Hitzler, Wright State University, USA, pascal@pascal-hitzler.de\\ Adila Krisnadhi, Universitas Indonesia, krisnadhi@gmail.com}
\date{August 2018}

\maketitle


We provide a detailed example for modular ontology modeling based on ontology design patterns. It is similar to the Chess Ontology tutorial in \cite{chess-odpbook}, which we suggest to read first. We will be less verbose in this tutorial; we provide it because additional examples should be helpful for those interested in adopting the modular ontology modeling methodology -- see \cite{chess-odpbook} and the book \cite{odp-book} in which it is contained. 

We assume that the reader is familiar with the Web Ontology Language OWL \cite{FOST,owl2-primer}.

Before we dive into the actual modeling, let us present the general workflow which we recommend for ontology modeling, and which is the same as in \cite{chess-odpbook}. The steps of this workflow are laid out in Figure \ref{fig:workflow}. We will refer to these steps, and explain them in more detail, as we advance through the tutorial.

\begin{figure}[t]
\begin{enumerate}
\item Define use case or scope of use cases.

\item Make competency questions while looking at possible data sources and scoping the problem, i.e., decide on what should be modeled now, and what should be left for a possible later extension. 

\item Identify key notions from the data and the use case and identify which pattern should be used for each. Many can remain ``stubs'' if detailed modeling is not yet necessary. Instantiate these key notions from the pattern templates, and adapt/change the result as needed. Add axioms for each module, informed by the pattern axioms. As a result of this step, we arrive at a set of modules for the final ontology.

\item Put the modules together and add axioms which involve several modules. Reflect on all class, property and individual names and possibly improve them. Also check module axioms whether they are still appropriate after putting all modules together.

\item Create OWL files.
\end{enumerate}
\caption{Ontology modeling workflow followed.}\label{fig:workflow}
\end{figure}

\section{Use Case}\label{sec:rec-usecase}

\emph{Step 1: Define use case or scope of use cases.}

Every ontology is designed for a purpose; this purpose may be defined by a use case, or by a set of use cases, or possibly by a set of potential use cases, which may include the future extensions or refinements of the ontology, and future reuse of the ontology by others. 

How specific should a use case be? Conventional wisdom may suggest that it is always better to be more specific. However, in the context of ontology modeling the case is not as clear-cut. A very specific use case may give rise to an ontology which is very specialized, i.e. modeling choices (so-called \emph{ontological commitments}) may be made which fit only the very specific and detailed use case. As a consequence, later modifications, e.g. by widening the scope of the application (and therefore of the underlying ontology) become very cumbersome as they may conflict with ontological commitments made earlier. 

Let us look at a very simple example of this. Say, our application involves movies and actors from the casts of these movies. It may first seem as if actor names could simply be attached to the movies using an OWL datatype property \textsf{hasActor}. E.g., this could be written in RDF Turtle as
\begin{verbatim}
:myMovie     :hasActor    "JaneSmith"  .
\end{verbatim}
This will be sufficient, e.g., if only the name of an actor is relevant for an application. 

However, it is conceivable that the application (and thus the ontology) may later on be extended in order to be able to list all movies in which a given actor was a cast member. Since it is likely that there may be different persons with the same name, such as Jane Smith, we would need to be able to identify which name strings identify the same, and which identify different actors, i.e., we have to disambiguate the name strings. Furthermore, Jane Smith may also be listed as actor under a different name, say Jane W. Smith. 

Using an OWL datatype property as above, however, was a modeling choice which prevents this. What we would need is URIs for actors. With this case our example would look like the following.
\begin{verbatim}
:myMovie       :hasActor   :janeSmith1 .
:janeSmith1    :hasName    "JaneSmith" .
\end{verbatim}

Further extensions of the application (or attempted reuses of the ontology), however, may pose yet additional problems. E.g., it may be desired to also list the character played by an actor in a specific movie. Given our current modeling choices, however, it seems unclear where to attach this information: If it is attached to the movie, then we would no longer be able to say which character in the movie was played by which actor. If we attach the information to the person, then we would no longer know which movie the character appeared in. If we attach it to both movie and person, then we would run into difficulties if characters appear in different movies, played by different actors. 

The solution in this case is to create another node in the graph, which stands for the \emph{actor role}. Our example would then look as thus.
\begin{verbatim}
:myMovie          :hasActor    :myMovieMissXRole .
:myMovieMissXRole :assumedBy   :janeSmith1 ;
                  :asCharacter :MissX .
:janeSmith1       :hasName     "JaneSmith" .
\end{verbatim}

We understand that we can make modeling choices which make future reuse easier, e.g., by making sure that we include enough nodes in the graph. This, of course, begs the question where to stop? If we follow this principle, then won't we end up with much too many nodes, blowing up the graphs?

This is a valid concern, of course, and there are not straightforward solutions for this issue which work in all circumstances. Generally speaking, we should strive for a balance, i.e., finding a soft spot somewhere between the extremes. Our approach using ontology design patterns addresses the issue as we will be able to reuse patterns which have been created and vetted by the community, and which provide a good trade-off between the extremes in many circumstances.

\begin{figure}[t]
\centerline{\includegraphics[width=.95\textwidth]{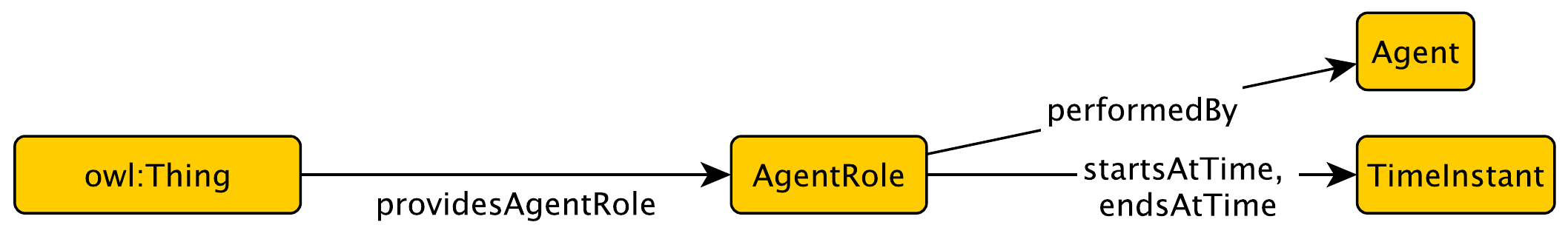}}
\caption{Generic AgentRole pattern}\label{fig:rec-agentrole}
\end{figure}

\begin{figure}[t]
\centerline{\includegraphics[width=.5\textwidth]{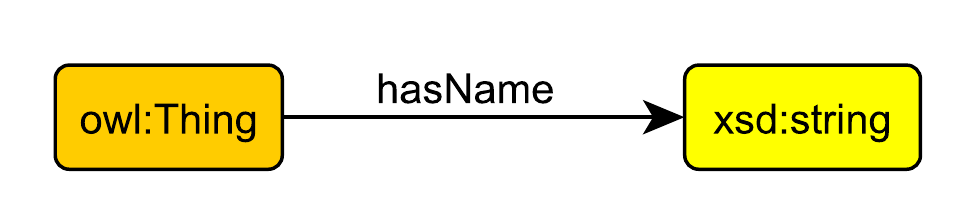}}
\caption{Generic NameStub pattern}\label{fig:rec-namestub}
\end{figure}

\begin{figure}[t]
\centerline{\includegraphics[width=.95\textwidth]{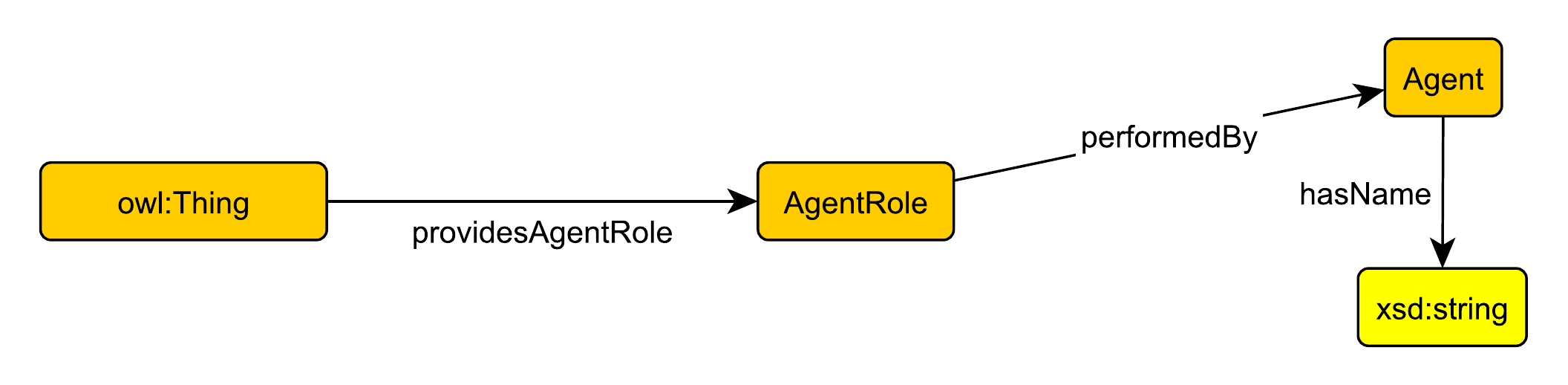}}
\caption{Joining the AgentRole and NameStub patterns}\label{fig:rec-moviegen}
\end{figure}

\begin{figure}[t]
\centerline{\includegraphics[width=.95\textwidth]{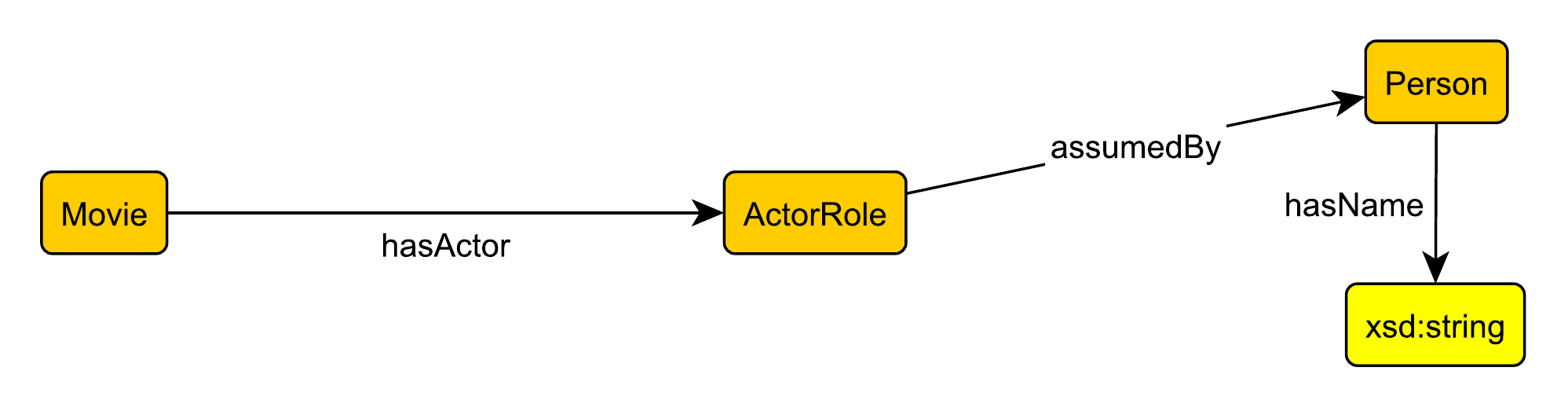}}
\caption{The movie snippet: using Figure \ref{fig:rec-moviegen} as a template}\label{fig:rec-movie}
\end{figure}

Returning to the movie example, there are two patterns which would be the standard choices in this situation, the AgentRole and the NameStub pattern. Class diagrams for these two are shown in Figures \ref{fig:rec-agentrole} and \ref{fig:rec-namestub}. We can now take the AgentRole pattern and remove the TimeInstant class, and join it with the NameStub pattern by using Agent instead of owl:Thing, see Figure \ref{fig:rec-moviegen}. Finally, we use the resulting class diagram as a template and make renamings of class and property names to fit it to our more specific use case; the result is displayed in Figure \ref{fig:rec-movie}. This process exemplifies our intended use of ontology design patterns: We use them as templates and specialize and join them in order to obtain a draft of our desired model.

After this discussion, let us now return to the actual task at hand, namely to define a use case scenario for our worked example. The setting we have in mind concerns online cooking recepies, and in particular the task of integrating recipes from different websites in order to enable a fine-grained cross-website search for recipes:

\emph{Design an ontology which can be used as part of a ``recipe discovery'' website. The ontology shall be set up such that content from existing recipe websites can in principle be mapped to it (i.e., the ontology gets populated with data from the recipe websites). On the discovery website, detailed graph-queries (using the ontology) shall produce links to recipes from different recipe websites as results. The ontology should be extendable towards incorporation of additional external data, e.g., nutritional information about ingredients or detailed information about cooking equipment.}

Let us make a few remarks about the scenario we have just defined. First of all, we notice that data will come from multiple sources which are not exactly specified. This means that our ontology needs to be general enough to accomodate different conceptual representations on the source side. Second, the ontology shall be extendable towards additional related data, meaning that we have to accomodate such extension capabilities, to a reasonable extent, without knowing what these future extensions would exactly look like, and this again asks for a rather general model. Third, fine-grained search for recipes shall be possible, meaning that our ontology needs to be specific enough to allow these. The scenario thus calls for a reasonable trade-off between specifity and generality, i.e., it is a typical use-case for ontology design pattern based modular ontology modeling.

\section{Competency Questions and Data Sources}\label{sec:rec-competency}

\emph{Step 2: Make competency questions while looking at possible data sources and scoping the problem, i.e., decide on what should be modeled now, and what should be left for a possible later extension.}

Competency questions are queries, formulated in natural language, which could potentially used for retrieval of data from the knowledge base. They help to further specify the use cases, i.e., main classes of potential queries should be represented. 

For our scenario, possible competency questions are the following. 
\begin{enumerate}
\item Gluten-free low-calorie desserts.
\item How do I make a low-carb pot roast?
\item How do I make a Chili without beans?
\item Sweet breakfast under 100 calories.
\item Breakfast dishes which can be prepared quickly with 2 potatoes, an egg, and some flour.
\item How do I prepare Chicken thighs in a slow cooker?
\item A simple recipe with pork shoulder and spring onions.
\item A side prepared using Brussels sprouts, bacon, and chestnuts.
\end{enumerate}

The competency questions already indicate some parameters that will be important, e.g.:
\begin{itemize}
\item Retrieval of cooking instructions.
\item Search by ingredients.
\item Search by properties of the prepared food, e.g. calorie or carb content. 
\item Search by properties such as cooking time, simplicity.
\end{itemize}

At this stage, at the latest, it is also necessary to look at possible data sources. A quick Web search provides a significant number of recepie websites, e.g., allrecipes.com, food.com, epicurious.com. Pages commonly list ingredients, cooking instructions, and sometimes other information such as nutritional information. Additional nutritional information is, e.g., available from Google Knowledge Graph nutrition data.\footnote{\url{https://search.googleblog.com/2013/05/time-to-back-away-from-cookie-jar.html}} Data is usually not available in structured form, i.e., for an application it will be necessary to extract content from text-based web pages, which can be a tricky task in itself; however herein we only concern ourselves with producing a suitable underlying ontology. 

Looking at the data sources now prompts us to go back to the competency questions, to reevaluate them. Some competency questions may have to be dropped or modified at first, e.g., recipe websites seem to rarely mention equipment, like slow cookers, separately, identify a breakfast as \emph{sweet}, or use  classification such as low-carb or gluten-free. We should keep these in mind, though, and make sure that the ontology we produce is extendable towards future inclusion of such aspects. At the same time, inspection of the data may yield further insights regarding data that could now or in the future be included, such as recipe authors, peer recommendations, cooking time, level of difficulty, or category tags such as \emph{dessert} or \emph{side}. These can either be incorporated right away, or alternatively extensibility towards future inclusion can be kept in mind during modeling.

The decision process regarding what to model now versus later is called \emph{scoping}. At the end of this step, we should have arrived at a clear idea concerning the scope of the target ontology. 

\section{Key Notions to Modules}\label{sec:rec-notions}

\emph{Step 3: Identify key notions from the data and the use case and identify which pattern should be used for each. Many can remain ``stubs'' if detailed modeling is not yet necessary. Instantiate these key notions from the pattern templates, and adapt/change the result as needed. Add axioms for each module, informed by the pattern axioms. As a result of this step, we arrive at a set of modules for the final ontology.}

Think of the key notions as the main classes of things appearing in the competency questions or which you identify from the data sources. Obvious possible key notions which come to mind are recipe, food, time, equipment, classification of food prepared (e.g., as side), difficulty level, nutritional information. Let's go through these one by one and refine the list while creating corresponding schema diagrams. After that, we will talk about axiomatizing them.

\subsection{Class Diagrams}

\subsection*{Recipe}

Recipe is an obvious candidate for a class, it is central to what we intend to do, and in addition we may want to notice already that the name of the recipe, which is often identical with the food which is going to be prepared, should be recorded. For the latter, we should probably use the NameStub. We now also want to identify a pattern which will be the basis for the core of the recipe modeling. 

Of course, we have not yet discussed other patterns than NameStub and AgentRole.  One way to approach this is to go through a list of known patterns\footnote{\url{htt://www.ontologydesignpatterns.org} sports many patterns, however they and their documentations are of very differing quality.} and to contemplate which may fit best, and sometimes there seem to be more than one candidate. Let us look at three more or less obvious candidates. 

Let us first check whether it makes sense to think of recipes as documents. There is certainly a perspective from this this seems valid: in the end, isn't it simply a document which we retrieve from the Web when we download a recipe? However, document seems to be a rather generic notion which does not naturally cater for key aspects of a recipe such as having ingredients, or taking a particular time. We also wouldn't say about a document whether it's low-carb or not. This line of thinking may lead us to the conclusion that a document may \emph{contain} a recipe \emph{description}, but that a recipe as such is a different type of entity. 

Since recipes usually contain step-by-step descriptions of the food preparation process, another alternative may be to think of a recipe as a sequence, which is another fundamental ontology design pattern. But then it also seems clear that many of the aspects important for our competency questions are not naturally catered for by the notion of sequence, e.g., what are ingredients in relation to recipe as a sequence? This line of thinking may lead us to the conclusion that some parts of the recipe -- the cooking steps -- may be representable as a sequence, but the whole of the recipe is much more than that. On the other hand, our competency questions do not indicate that the preparation steps sequence as such is particularly relevant to our task, namely the discovery of recipes.

We could also think of recipes as \emph{processes} which may help us to emphasize input and output aspects. This may indeed be a valuable perspective. However, the notion of process may usually allude to much more rigid and well-defined sequences of actions, so we would have to have a very detailed look at a process ontology design pattern to decide whether it provides the right perspective for our purpose.

The perspective we will actually take here is that a recipe is a type of description. Indeed, the general description pattern \cite{GangemiM03} has a specialization to plans, and indeed it seems a reasonable perspective to think of recipes as plans (to produce something). 

\begin{figure}[t]
\centerline{\includegraphics[width=.8\textwidth]{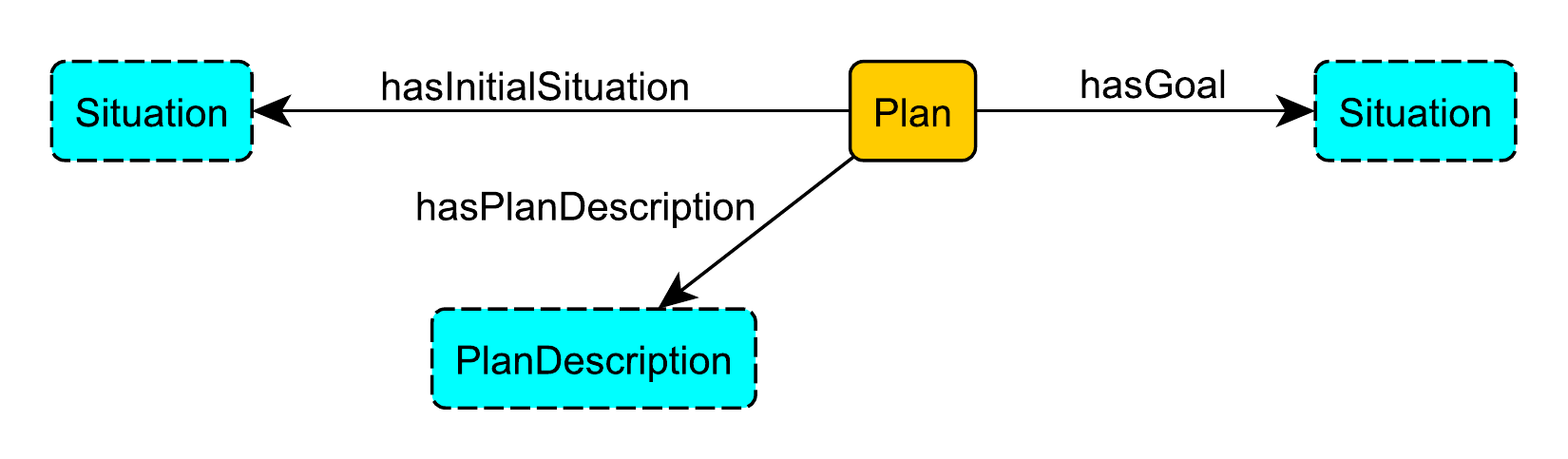}}
\caption{Basic Plan ontology design pattern: schema diagram. The dashed boxes indicate complex notions which would easily merit a pattern description in their own right.}\label{fig:rec-plan}
\end{figure}

\begin{figure}[t]
\centerline{\includegraphics[width=.8\textwidth]{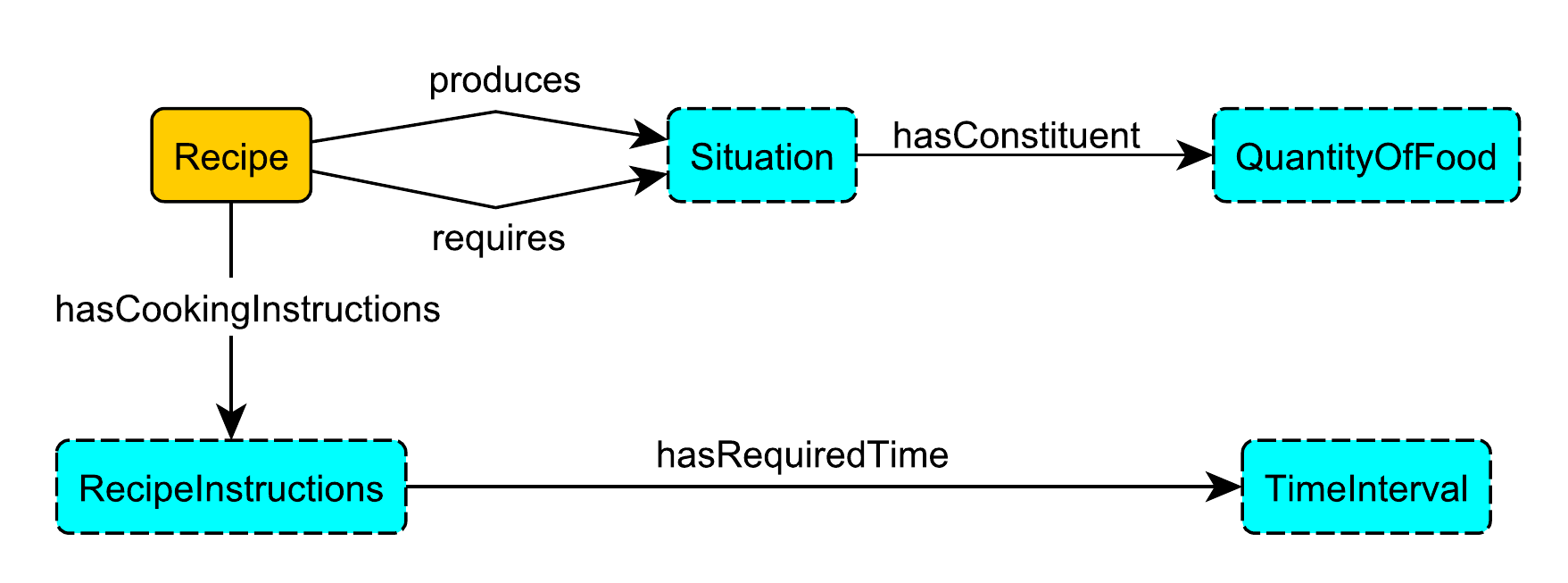}}
\caption{Recipe as plan}\label{fig:rec-recipeAsPlan}
\end{figure}

Let us look at a part of the Plan pattern which is depicted in Figure \ref{fig:rec-plan}. A plan leads from an initial situation to a situation which is understood as the goal of the plan. The initial situation would be one in which required ingredients (and equipment etc) would be available, while in the goal situation the prepared food would be available. In fact, the required ingredients, equipment etc. are necessary for these respective situations, i.e. they are \emph{constituents} for them. Putting these thoughts together, we can arrive at a first piece of the Recipe module, as an instantiation of the Plan template. Its schema diagram is depicted in Figure \ref{fig:rec-recipeAsPlan}. There is much more to be said about descriptions, plans, situations, etc., but we will not go into detail here. See \cite{GangemiM03} for a central reference.


\subsection*{Food}

Proceeding with our list of potential key notions, the next one is \emph{food}. This seems a little unspecific. What exactly is meant by this? Well, food can be things like cucumbers, potatoes, eggs, lasagna, and Chicken Kiev. Perhaps we should distinguish between ingredients and results, i.e., full dishes? But wait a second, what about, say, Pesto Genovese? It's not a dish by itself, but an ingredient in some recipes; yet there are also recipes how to make Pesto Genovese. Indeed, many cooking ingredients are already processed from even more basic ingredients. So it probably will not make much sense to try to distinguish between ingredients and dishes when talking about recipes.\footnote{The type of discussion exemplified in this paragraph is central to coming up with good key notions and modules. It is extremely helpful to have this discussion in a group, as others are often so much better in finding flaws in our ideas than we are ourselves.}

But, if we say Pesto Genovese, what exactly do we mean? Do we mean Pesto Genovese in general, as such, or do we mean, say, two teaspoons of it, as required by some recipe? Indeed it seems that for recipes the quantity of a required or produced food item is also important. 

We're starting to narrow this down. \emph{Quantity} seems like an ontology design pattern which we should make use of for our purposes, and ``a quantity of food'' seems to be a central concept for modeling recipes as plans, to be used both on the input and on the output side of the recipe as plan.

\begin{figure}[t]
\centerline{\includegraphics[width=.99\textwidth]{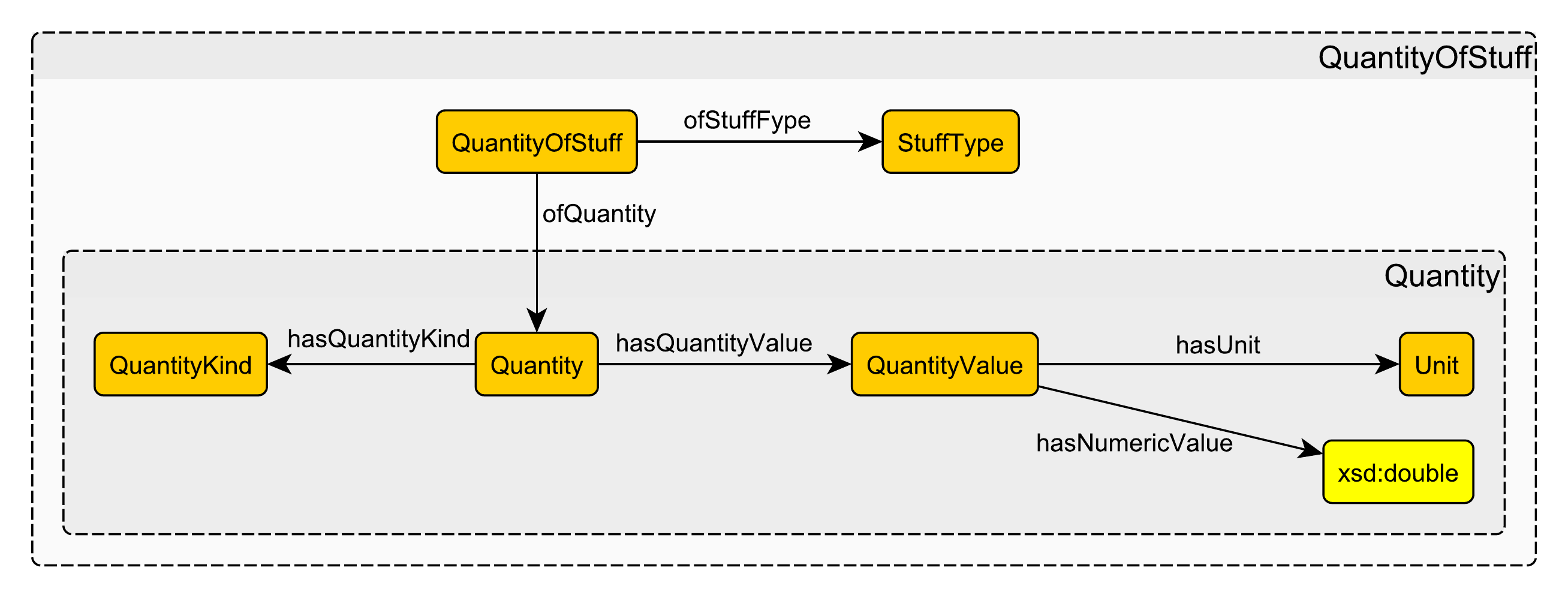}}
\caption{The QuantityOfStuff pattern, the inner box indicates the Quantity pattern. There is much more to be said about quantities, but we will not further dwell on this here}\label{fig:rec-quantity}
\end{figure}

\begin{figure}[t]
\centerline{\includegraphics[width=.99\textwidth]{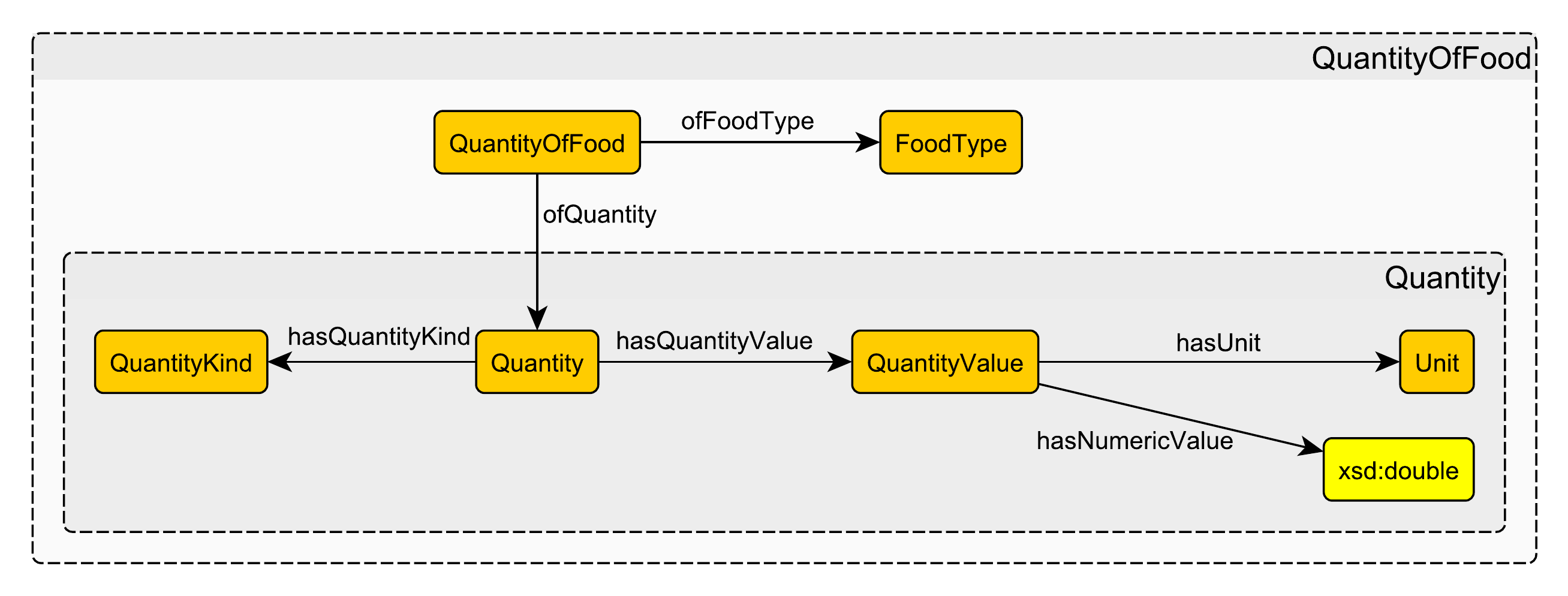}}
\caption{QuantityOfFood module, as an instance of the Quantity pattern from Figure \ref{fig:rec-quantity}.}\label{fig:rec-quantityOfFood}
\end{figure}

So we understandt hat there should be a concept of QuantityOfFood (like, 2 tsp of Pesto) which is always of some quantity (like, 2 tsp) and at the same time is of some type of foodstuff (say, Pesto). The foodstuff can thus be understood as a FoodType (like, Pesto, or potato), namely the type of stuff the quantity of food consists of. See Figures \ref{fig:rec-quantity} and \ref{fig:rec-quantityOfFood} for schema diagrams. Our pattern for Quantity is very much directly derived from QUDT.\footnote{\url{http://qudt.org/}}

\subsection*{Equipment, Classification, Difficulty Level}

Next on our list of key notions is \emph{equipment}, such as slow cooker, blender, etc. However, while keeping track of the occasional special equipment may be helpful, our scenario does not call for a detailed modeling of kitchen equipments at this stage. So let us decide to delay such detailed modeling for the moment, i.e., we consciously restrict the scope of our model.

Decisions such as this are very important during the modeling process, as they limit the scope of the ontology. Indeed, it is impossible to always model all details, as we would end up with a model of almost everything in this case. At the same time, however, we would like to keep in mind that our ontology may be reused later and possibly repurposed for a scenario in which detailed modeling of equipment may be more important. This means, that we do not want to simply introduce a datatype property such as \verb|requiresEquipment| with strings -- the names of the equipment -- as range. We rather want to utilize a slightly more sophisticated approach where we at least have a node as placeholder for the equipment entity.

\begin{figure}[t]
\centerline{\includegraphics[width=.99\textwidth]{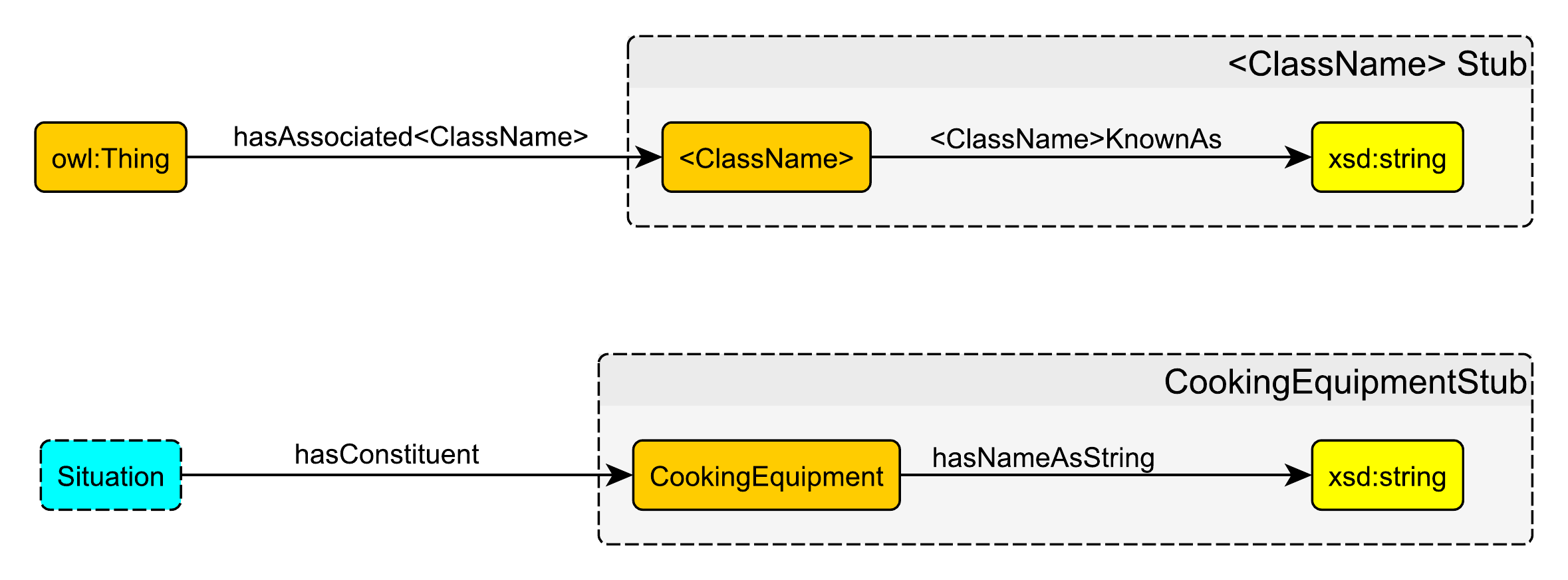}}
\caption{Top, the Stub (meta)pattern. Bottom, its instantiation for equipment.}\label{fig:rec-stub}
\end{figure}

The corresponding ontology design pattern is called a \emph{Stub} \cite{stub-metapattern}, and it is depicted in Figure \ref{fig:rec-stub} together with the instantiation for equipment which we will use. It is really essentially the same as the NameStub pattern introduced earlier, the only difference being that the identifying string is not necessarily a name of the thing identified.

Note also that we attach the cooking equipment as constituent to a situation, which seems to be its natural place.

\begin{figure}[t]
\centerline{\includegraphics[width=.85\textwidth]{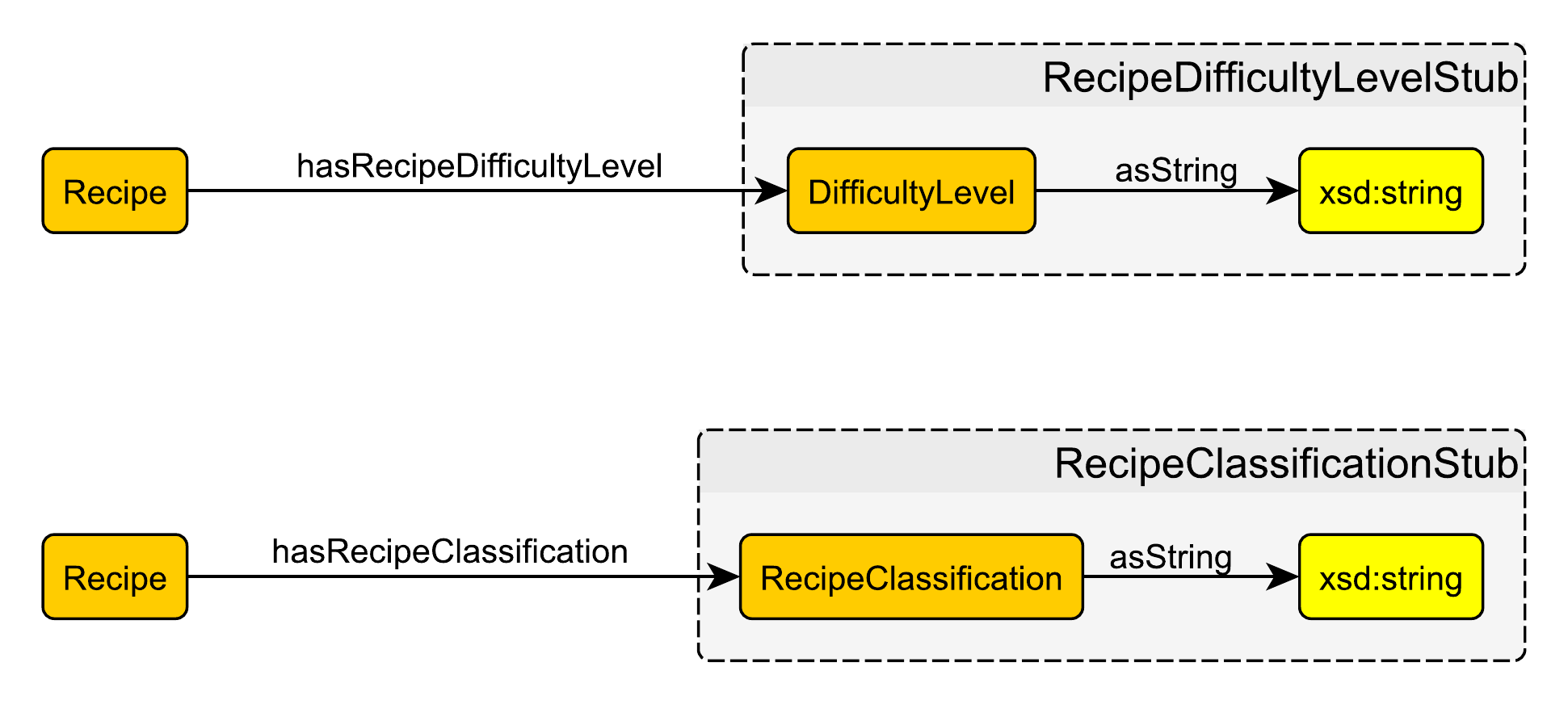}}
\caption{Stubs for DifficultyLevel and RecipeClassification.}\label{fig:rec-otherStubs}
\end{figure}

We opt for stubs also for other key notions we have identified, namely for DifficultyLevel and for RecipeClassification (such as low-carb, diabetic, etc.), i.e., for now the ontology will be able to hold only strings for these, but the model remains extendable if so desired in the future. The corresponding schema diagrams can be found in Figure \ref{fig:rec-otherStubs}.

We will use stubs also in other places, e.g., we have not further talked about FoodType as it appears in Figure \ref{fig:rec-quantityOfFood}. As before, it is conceivable that there may be a sophisticated model of different food types, but we will use a stub at this stage.

\subsection*{Nutritional Information}

\begin{figure}[t]
\centerline{\includegraphics[width=60mm]{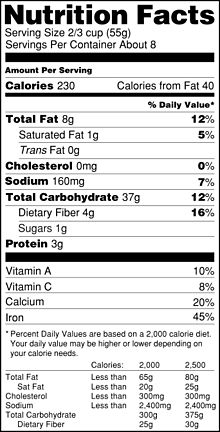}}
\caption{Example for a U.S. FDA Nutritional Facts label}\label{fig:rec-nfl}
\end{figure}

Next we turn our attention to \emph{nutritional information}, the next keyword on our list, and we opt to model this in somewhat more detail. More precisely, we will model the contents of Nutritional Facts labels as mandated in the U.S.A. for most food products,\footnote{see \url{https://en.wikipedia.org/wiki/Nutrition_facts_label\#United_States}} see Figure \ref{fig:rec-nfl} While this may seem overly specific, by virtue of our modular modeling approach it would be easy to replace the NutritionalInformation module with one tailored to, e.g., other countries, or other nutritional convictions. In fact, we will highlight this by creating a class \verb|US-Nutrition-Label| as a subclass of the generic NutritionalInformation class. 

\begin{figure}[t]
\centerline{\includegraphics[width=.99\textwidth]{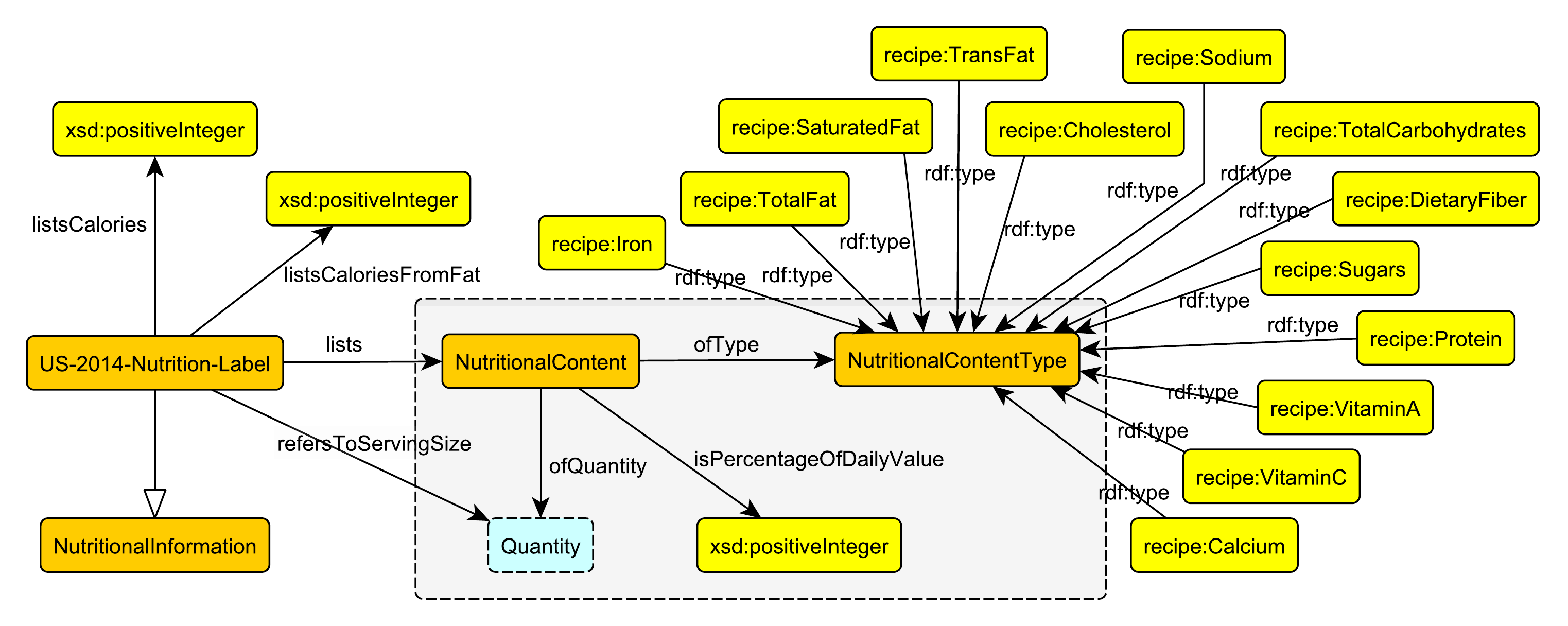}}
\caption{Nutritional Information module. The box indicates a modified instance of the QuantityOfStuff pattern.}\label{fig:rec-nutrinfo}
\end{figure}

These Nutritional Facts labels have highly structured content. We will of course not be concerned with layout issues, and it is also not necessary that we model all  content. E.g., we will not list ``\% Daily Value'' amounts for fat or sodium. We will list absolute amounts for Fat, Saturated Fat, Trans Fat, Cholesterol, Sodium, Carbs, Dietary Fiber, Sugars, and Protein, and ``\% Daily Value'' amounts for Vitamins A and C, Calcium, and Iron, which we represent as instances of a class \verb|NutritionalContentType|.\footnote{Essentially, we are creating a small \emph{controlled vocabulary} for substances of nutritional importance.} It seems obvious that we will reuse the \emph{QuantityOfStuff} pattern again, however as a percentage value is not really a quantity, we add an alternative to giving the quantity, which consists simply of a datatype property \verb|isPercentageOfDailyValue| with range \verb|xsd:positiveInteger|. See the right of Figure \ref{fig:rec-nutrinfo}.

Of course we also need to record the serving size to which the nutritional information refers, and this can again be done using the \emph{Quantity} pattern. We also list calorie content and calories-from-fat content as indicated in the figure.

\subsection*{Provenance}

We have worked through our list of keywords, but before we move on, let us briefly reflect whether there is anything else that needs modeling, which we can derive from our scenario description. And indeed, our scenario states that queries shall produce links to recipes from different recipe websites -- however, or modeling so far did not include anything which would make it possible to track where a recipe came from. We thus need to do some provenance modeling. 

\begin{figure}[t]
\centerline{\includegraphics[width=.6\textwidth]{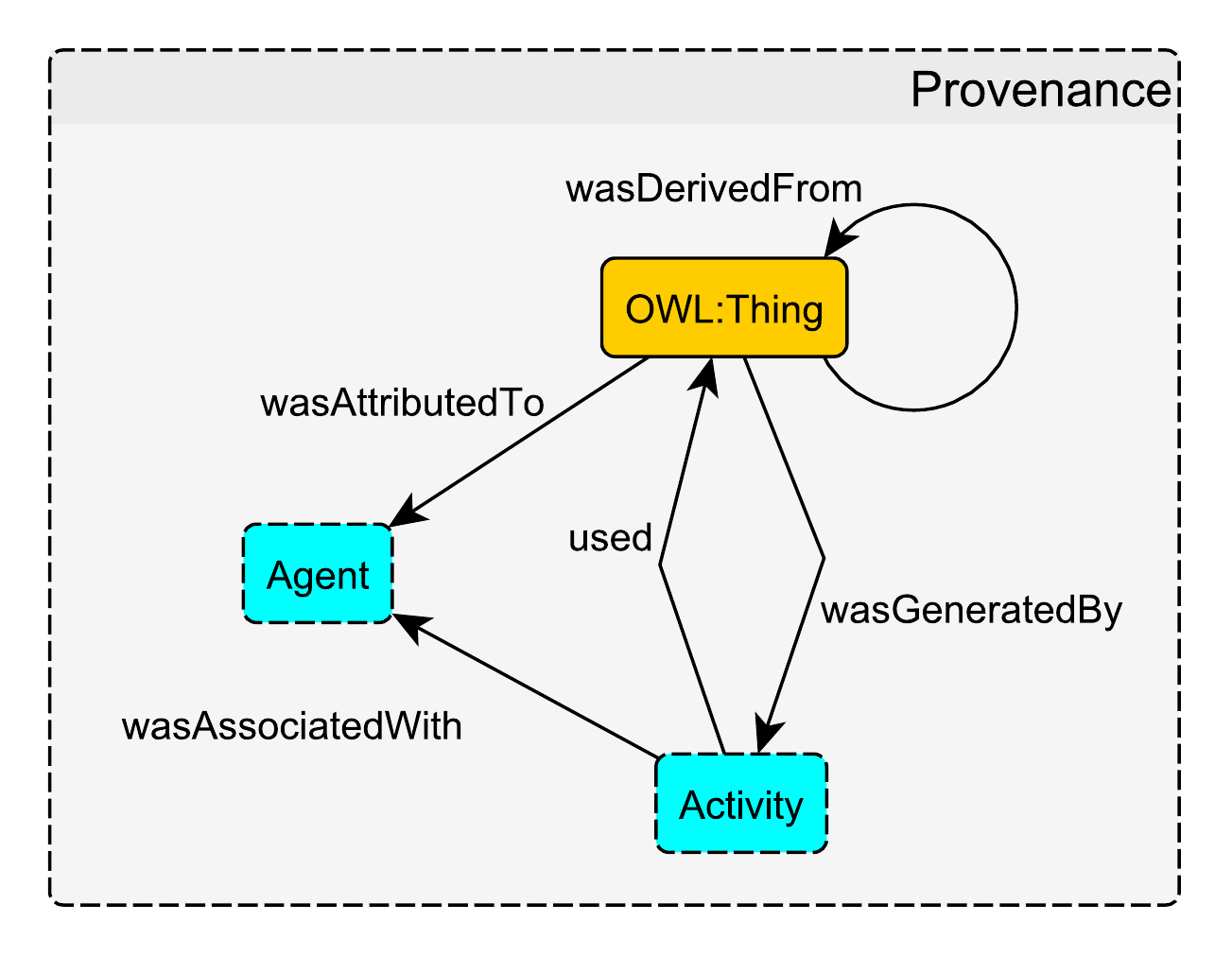}}
\caption{Provenance pattern}\label{fig:rec-provenance}
\end{figure}

The schema diagram of a generic provenance pattern, as derived from PROV-O \cite{provo} and mentioned in \cite{partof-pattern}, is provided in Figure \ref{fig:rec-provenance}. The key idea of this is that everything (any \verb|owl:Thing|) for which provenance is important, was generated by some activity (in our case, web retrieval) which used some other thing (in our case, a recipe website). The item under consideration (the recipe) may also be directly related to its origin (the recipe website) using the property \verb|wasDerivedFrom|. In addition, agents may be involved in activities.

\begin{figure}[t]
\centerline{\includegraphics[width=.8\textwidth]{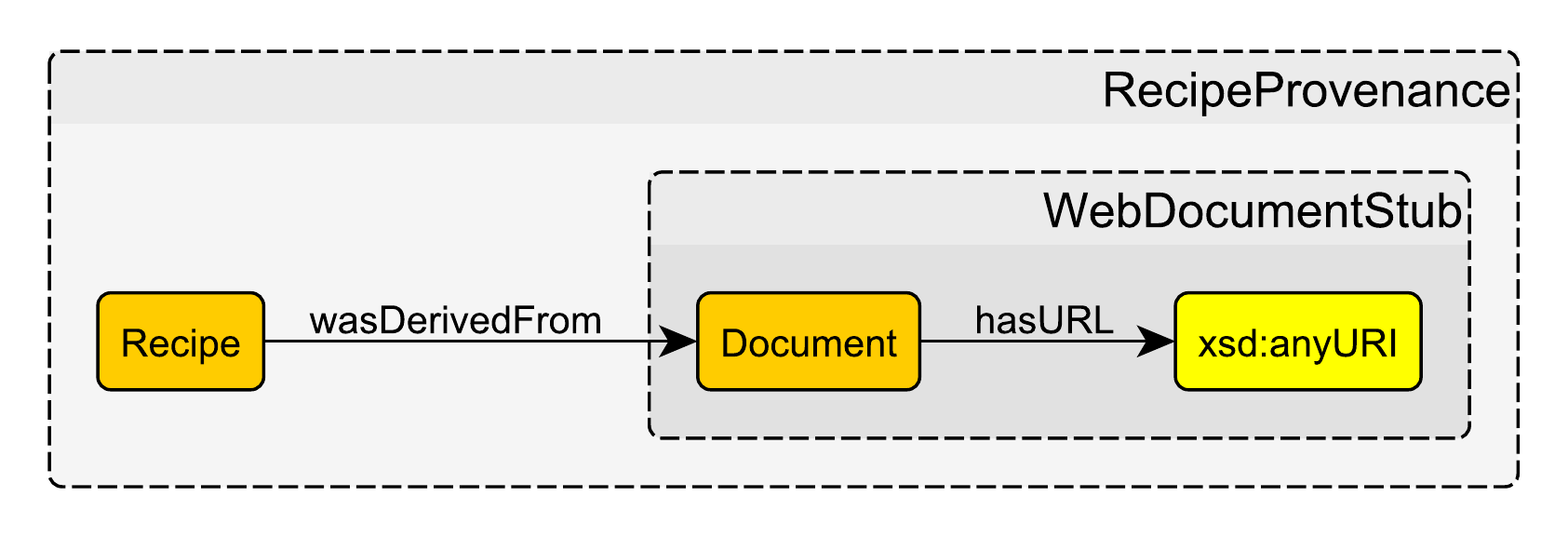}}
\caption{Recipe provenance module}\label{fig:rec-recipe-provenance}
\end{figure}

For our purpose, it will suffice to reuse a small part of this pattern, namely the \verb|wasDerivedFrom| property. Derivation in this case is from a document which has a URL (i.e., a website), and we can use a Document stub for this. The resulting module is depicted in Figure \ref{fig:rec-recipe-provenance} 

\subsection{Axiomatizations}

\begin{figure}[t]
\begin{tabbing}
RecipeClassificationxxx	\=Stub (as starter, dessert, etc)\kill
Recipe 			\>Plan\\
RecipeName		\>NameStub\\
RecipeInstructions	\>Document\\
TimeInterval		\>temporal information\\
QuantityOfFood          \>QuantityOfStuff\\
Quantity		\>Quantity\\
Equipment		\>Stub\\
FoodType		\>Stub\\
Difficultylevel		\>Stub\\
RecipeClassification	\>Stub\\
NutritionalInfo		\>unspecified pattern using QuantityOfStuff\\
Source 			\>Provenance
\end{tabbing}
\caption{All key notions together with corresponding patterns used}\label{fig:rec-notions}
\end{figure}

We have now produced diagrams for all key notions we had identified, and have used schema diagrams of general ontology design patterns to produce them. The list of key notions, together with the used patterns can be found in Figure~\ref{fig:rec-notions}.

We now turn to producing OWL axioms for all modules. We use the earlier schema diagrams as guidance. Usually, axioms would be derived from the axioms provided with the patterns, but we will recreate them from scratch, in order to gain a deeper understanding of them. We will in fact produce a rather exhaustive list of axioms which seem appropriate for our model, while steering away from overly strong ontological commitments.

\begin{figure}[t]
\centerline{\includegraphics[width=.3\textwidth]{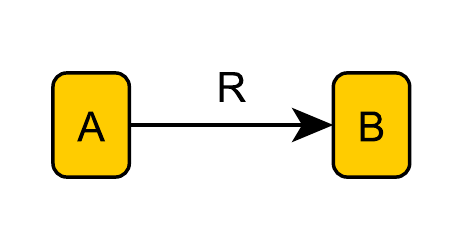}}
\caption{Generic node-edge-node schema diagram for explaining systematic axiomatization}\label{fig:rec-ARB}
\end{figure}

There is a systematic way to look at each node-edge-node triple in the schema diagram in order to decide which axioms should be added: Given a node-edge-node triple with nodes $A$ and $B$ and edge $R$ from $A$ to $B$, as depicted in Figure \ref{fig:rec-ARB}, we check all of the following axioms whether they should be included.\footnote{The OWLAx Prot\'eg\'e plug-in \cite{SarkerKH16} provides a convenient interface for adding these axioms.} We list them in natural language, see Figure \ref{fig:generic-triple-axioms-DL} for the formal versions in description logic notation, and Figure \ref{fig:generic-triple-axioms-Manchester} for the same in Manchester syntax, where we also list our names for these axioms.
\begin{compactenum}
\item $A$ and $B$ are disjoint.
\item The domain of $R$ is $A$.
\item For every $B$ which has an inverse $R$-filler, this inverse $R$-filler is in $A$. In other words, the domain of $R$ scoped with $B$ is $A$.
\item The range of $R$ is $B$.
\item For every $A$ which has an $R$-filler, this $R$-filler is in $B$. In other words, the range of $R$ scoped with $A$ is $B$.
\item For every $A$ there has to be an $R$-filler in $B$.
\item For every $B$ there has to be an inverse $R$-filler in $A$.
\item $R$ is functional.
\item $R$ has at most one filler in $B$.
\item For every $A$ there is at most one $R$-filler.
\item For every $A$ there is at most one $R$-filler in $B$.
\item $R$ is inverse functional.
\item $R$ has at most one inverse filler in $A$.
\item For every $B$ there is at most one inverse $R$-filler.
\item For every $B$ there is at most one inverse $R$-filler in $A$.
\end{compactenum}

Domain and range axoims are items 2--5 in this list. Items 6 and 7 are extistential axioms. Items 8--15 are about variants of functionality and inverse functionality. All axiom types except disjointness and those utilizing inverses also apply to datatype properties.

\begin{figure}[t]
\begin{center}
\begin{minipage}{.3\textwidth}
\begin{compactenum}
\item $A\sqcap B\sqsubseteq \bot$
\item $\exists R.\top \sqsubseteq A$
\item $\exists R.B\sqsubseteq A$
\item $\top \sqsubseteq \forall R.B$ 
\item $A\sqsubseteq \forall R.B$
\end{compactenum}
\end{minipage}
\begin{minipage}{.3\textwidth}
\begin{compactenum}
\setcounter{enumi}{5}
\item $A\sqsubseteq R.B$
\item $B\sqsubseteq R^-.A$
\item $\top \sqsubseteq \mathord{\leq} 1 R.\top$ 
\item $\top \sqsubseteq \mathord{\leq} 1 R.B$
\item $A\sqsubseteq \mathord{\leq} 1 R.\top$
\end{compactenum}
\end{minipage}
\begin{minipage}{.3\textwidth}
\begin{compactenum}
\setcounter{enumi}{10}
\item $A\sqsubseteq \mathord{\leq} 1 R.B$
\item $\top \sqsubseteq \mathord{\leq} 1 R^-.\top$ 
\item $\top \sqsubseteq \mathord{\leq} 1 R^-.A$
\item $B \sqsubseteq \mathord{\leq} 1 R^-.\top$ 
\item $B \sqsubseteq \mathord{\leq} 1 R^-.A$
\end{compactenum}
\end{minipage}
\caption{Most common axioms which could be produced from a single edge $R$ between nodes $A$ and $B$ in a schema diagram: description logic notation.}\label{fig:generic-triple-axioms-DL}
\end{center}
\end{figure}

\begin{figure}[t]
\begin{compactenum}
\item $A$ \textsf{DisjointWith} $B$ \hfill (disjointness)
\item $R$ \textsf{some} \texttt{owl:Thing} \textsf{SubClassOf} $A$ \hfill (domain)
\item $R$ \textsf{some} $B$ \textsf{SubClassOf} $A$ \hfill (scoped domain)
\item \texttt{owl:Thing} \textsf{SubClassOf} $R$ \textsf{only} $B$ \hfill (range)
\item $A$ \textsf{SubClassOf} $R$ \textsf{only} $B$ \hfill (scoped range)
\item $A$ \textsf{SubClassOf} $R$ \textsf{some} $B$ \hfill (existential)
\item $B$ \textsf{SubClassOf inverse} $R$ \textsf{some} $A$ \hfill (inverse existential)
\item \texttt{owl:Thing} \textsf{SubClassOf} $R$ \textsf{max} $1$ \texttt{owl:Thing} \hfill (functionality)
\item \texttt{owl:Thing} \textsf{SubClassOf} $R$ \textsf{max} $1$ $B$ \hfill (qualified functionality)
\item $A$ \textsf{SubClassOf} $R$ \textsf{max} $1$ \texttt{owl:Thing} \hfill (scoped functionality)
\item $A$ \textsf{SubClassOf} $R$ \textsf{max} $1$ $B$ \hfill (qualified scoped functionality)
\item \texttt{owl:Thing} \textsf{SubClassOf inverse} $R$ \textsf{max} $1$ \texttt{owl:Thing} \hfill (inverse functionality)
\item \texttt{owl:Thing} \textsf{SubClassOf inverse} $R$ \textsf{max} $1$ $A$ \hfill (inverse qualified functionality)
\item $B$ \textsf{SubClassOf inverse} $R$ \textsf{max} $1$ \texttt{owl:Thing} \hfill (inverse scoped functionality)
\item $B$ \textsf{SubClassOf inverse} $R$ \textsf{max} $1$ $A$ \hfill (inverse qualified scoped functionality)
\end{compactenum}
\caption{Most common axioms which could be produced from a single edge $R$ between nodes $A$ and $B$ in a schema diagram: Manchester syntax.}\label{fig:generic-triple-axioms-Manchester}
\end{figure}

\subsection*{Recipe as Plan}

We now return to our recipe example, more precisely to Figure \ref{fig:rec-recipeAsPlan}. We will henceforth use Manchester syntax only; the description logic variants can be found in the appendix. The axioms can be found in Figure \ref{fig:rec-recipe-as-plan-axioms} Note that, generally speaking, we prefer scoped versions of axioms, as they represent the weaker axioms from the perspective of formal semantics.

\begin{figure}[t]
\begin{compactenum}
\item \texttt{Recipe} \textsf{SubClassOf} \texttt{requires} \textsf{only} \texttt{Situation}
\item \texttt{Recipe} \textsf{SubClassOf} \texttt{requires} \textsf{some} \texttt{Situation}
\item \texttt{Recipe} \textsf{SubClassOf} \texttt{produces} \textsf{only} \texttt{Situation}
\item \texttt{Recipe} \textsf{SubClassOf} \texttt{produces} \textsf{some} \texttt{Situation}
\item \texttt{hasCookingInstructions} \textsf{some} \texttt{RecipeInstructions} \textsf{SubClassOf} \texttt{Recipe}
\item \texttt{Recipe} \textsf{SubClassOf} \texttt{hasCookingInstructions} \textsf{only} \texttt{RecipeInstructions}
\item \texttt{Recipe} \textsf{SubClassOf} \texttt{hasCookingInstructions} \textsf{some} \texttt{RecipeInstructions}
\item \texttt{RecipeInstructions} \textsf{SubClassOf inverse} \texttt{hasCookingInstructions} \textsf{some} \texttt{Recipe}
\item \texttt{RecipeInstructions} \textsf{SubClassOf inverse} \texttt{hasCookingInstructions} \textsf{max} $1$ \texttt{Recipe}
\item \texttt{RecipeInstructions} \textsf{SubClassOf} \texttt{hasRequiredTime} \textsf{only} \texttt{TimeInterval}
\item \texttt{Recipe} \textsf{SubClassOf} \texttt{requires} \textsf{some} (\texttt{hasConstituent} \textsf{some} \texttt{QuantityOfFood})
\item \texttt{Recipe} \textsf{SubClassOf} \texttt{produces} \textsf{some} (\texttt{hasConstituent} \textsf{some} \texttt{QuantityOfFood})
\end{compactenum}
\caption{Axioms for Figure \ref{fig:rec-recipeAsPlan}}\label{fig:rec-recipe-as-plan-axioms}
\end{figure}

Items 1 and 2 refer to the \texttt{requires} edge and adjacent nodes in Figure \ref{fig:rec-recipeAsPlan}. Item 1 is a scoped range restriction; note that we do not specify a scoped domain, because we feel that a statement which says that quantities of food can only be ingredients in recipes (and in nothing else) may seem too restrictive. For a similar reason, we specify only one of the standard existential axioms, in Item 2. 
Items 3 and 4 are analogous, for the \texttt{produces} edge and adjacent nodes. Items 5--9 refer to the \texttt{hasCookingInstructions} edge and adjacent nodes; in this case we have scoped domain and scoped range expressions, both existentials, and a cardinality expression. The cardinality expression 9 states that every entity in the class RecipeInstructions can be associated as cooking instructions to at most one recipe. Item 10 is a scoped range expression for the \texttt{hasRequiredTime} edge and adjacent nodes; note that none of the other axioms seems fully appropriate in this case, e.g., other things can have required times as well. Items 11 and 12 are additional axioms which involve two properties and thus do not come from the list in Figure \ref{fig:generic-triple-axioms-Manchester}. They state that each recipe requires some \texttt{QuantityOfFood} to begin with, and also always produces some \texttt{QuantityOfFood}.

In short, we include the following axioms. For \texttt{requires}: scoped range, existential; for \texttt{produces}: scoped range, existential; for \texttt{has\-CookingIn\-structions}: scoped domain, scoped range, existential, inverse existential, inverse qualified scoped functionality; for \texttt{hasRequiredTime}: scoped range. We also have the additional axioms 11 and 12 from Figure \ref{fig:rec-recipe-as-plan-axioms}.

Furthermore, we declare disjointness axioms: \texttt{Recipe}, \texttt{QuantityOfFood}, \texttt{Situation}, \texttt{RecipeInstructions}, \texttt{TimeInterval} are mutually disjoint.

After going through the standard axiom candidates for each node-edge-node tripel, we also contemplate whether there should be any axioms spanning more nodes or edges. However, none such seem to be appropriate in this case.

\subsection*{QuantityOfFood}

We refer to Figure \ref{fig:rec-quantityOfFood}. Instead of listing formal axioms, we describe them by using the axiom names we have introduced in Figure \ref{fig:generic-triple-axioms-Manchester}. We thus have the following standard axioms. For \texttt{ofFoodType} and \texttt{ofQuantity}: scoped range, existential; for \texttt{hasQuantityKind} and \texttt{hasQuantityValue}: scoped domain, scoped range, existential, inverse existential, scoped qualified functionality; for \texttt{hasUnit}: scoped range, existential, scoped qualified functionality; for \texttt{hasNumericValue}: scoped range, existential, functionality. 

Furthermore, we declare disjointness axioms: \texttt{QuantityOfFood}, \texttt{Food\-Type}, \texttt{QuantityKind}, \texttt{Quantity}, \texttt{QuantityValue}, \texttt{Unit} are mutually disjoint. We do not add any other axioms.

There is more to be said about allowed units for each \texttt{QuantityKind}, but we will not dive into this here.

\subsection*{CookingEquipment, RecipeDifficultyLevel, RecipeClassification Stubs}

We refer to Figures \ref{fig:rec-stub} and \ref{fig:rec-otherStubs}. The axioms are as follows. For \texttt{hasConstituent}: existential; for \texttt{hasRecipe\-DifficultyLevel} and \texttt{has\-Re\-cipeClassification}: scoped domain, scoped range, existential, inverse existential; for \texttt{hasNameAsString} and \texttt{asString}: scoped range. 

Furthermore, we declare disjointness axioms: \texttt{Recipe}, \texttt{CookingEquip\-ment}, \texttt{DifficultyLevel}, \texttt{RecipeClassification} are mutually disjoint. We do not add any other axioms.

\subsection*{NutritionalInformation}

We refer to Figure \ref{fig:rec-nutrinfo}. The axioms are as follows. For \texttt{listsCalories}, for \texttt{listsCaloriesFromFat} and for \texttt{refersToServingSize}: scoped range, existential, functional; for \texttt{lists}: scoped domain, scoped range, existential; for \texttt{ofQuantity}, \texttt{ofType} and \texttt{isPercentageOfDailyValue}: scoped range, existential. 

Furthermore, we declare disjointness axioms: \texttt{US-2014-Nutrition-Label}, \texttt{NutritionalContent},  \texttt{Recipe}, \texttt{Quantity}, \texttt{NutritionalContent\-Type} are mutually disjoint. We do not add any other axioms.

\subsection*{RecipeProvenance}

We refer to Figure \ref{fig:rec-recipe-provenance}. The axioms are as follows. For \texttt{wasDerivedFrom}: scoped range, existential; for \texttt{hasURL}: scoped range. Furthermore, we declare disjointness axioms: \texttt{Recipe} and, \texttt{Document} are disjoint. We do not add any other axioms.

\section{Putting Things Together}\label{sec:rec-merge}

\emph{Step 4: Put the modules together and add axioms which involve several modules. Reflect on all class, property and individual names and possibly improve them. Also check module axioms whether they are still appropriate after putting all modules together.}

\begin{figure}[p]
\centerline{\includegraphics[width=.99\textwidth]{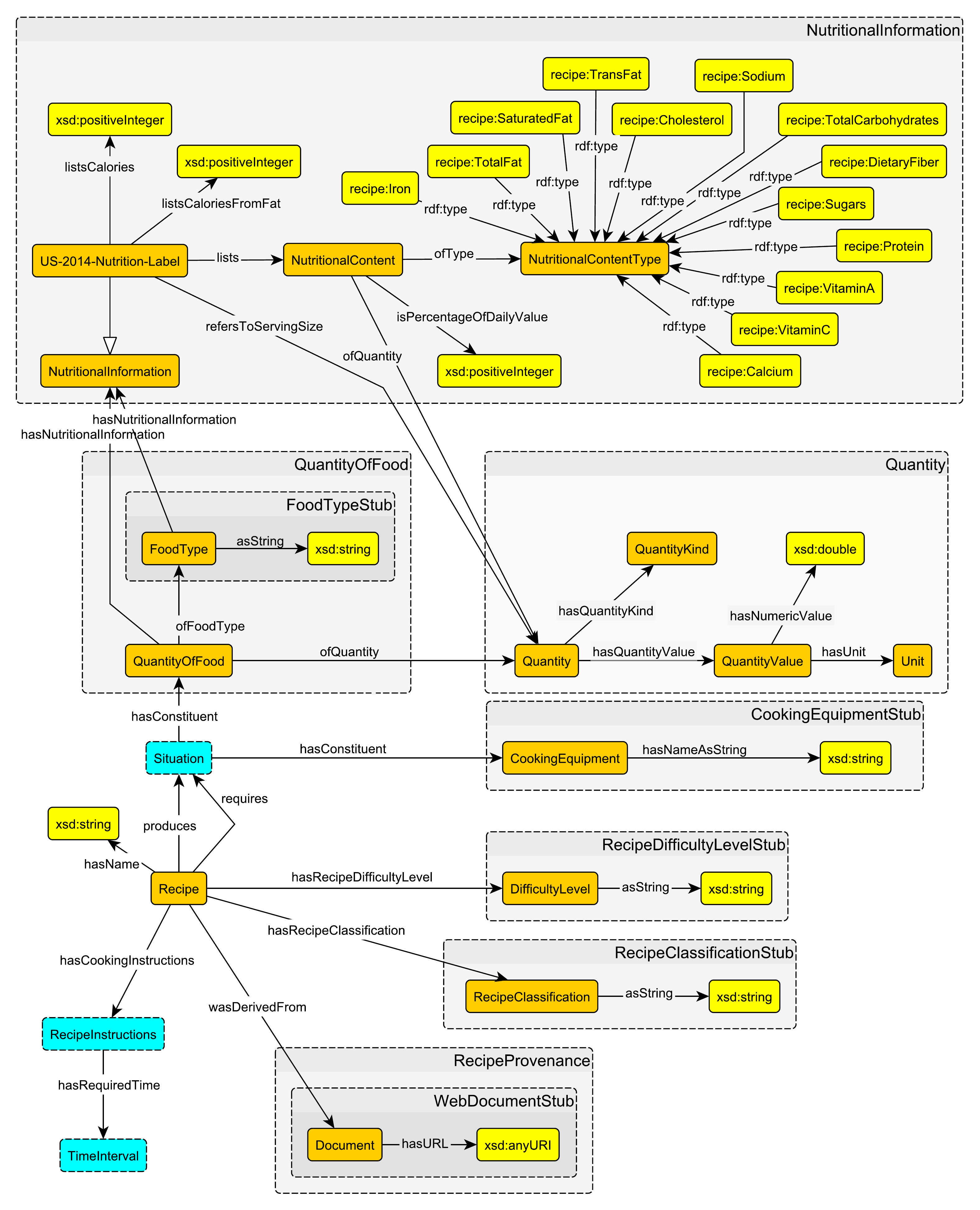}}
\caption{Recipe complete model}\label{fig:rec-recipe-all}
\end{figure}

We put the modules together by first joining the different schema diagram. The result is shown in Figure \ref{fig:rec-recipe-all}. The diagram also indicates most modules using grey boxes. We have already been very careful with proper naming, so in this case we do not have to make any corresponding improvmements. Also, all axioms remain appropriate even after joining.

We do add one datatype property, though, \texttt{hasName} as indicated in the diagram, to give names to recipes. While a name may not appear central at first sight, it should be easily retrievable (and it is often identical with the name of the QuantityOfFood produced by the recipe), and it should be helpful when displaying search results. We only decleare a scoped range for \texttt{hasName}. 
Finally, let us contemplate on additional axioms which we may want to have for the resulting ontology. When inspecting the diagram, additional axioms are sometimes indicated when the schema diagram, understood as an undirected graph, does not have a tree structure, i.e. contains cycles. There are several such cycles in the diagram, which we inspect carefully. However, it turns out that in each case, no additional axioms are warranted. E.g., several classes refer to \texttt{Quantity}, but there are no additional relationships between the different quantities to indicate. So we only need to add additional disjointness axioms, and in fact all classes depicted in the diagram are mutually disjoint.

Finally, we go back to the competency questions listed in Section \ref{sec:rec-competency}. We want to assess to what extent our ontology captures the required information to answer the competency questions. In cases where it does not, or not sufficiently, decisions need to be made whether the ontology should be modified or extended; but we will not go through this additional excercise herein.

The bulk of the competency questions concerns ingredients and equipment, which our ontology models. 

For the first question, we notice that desserts (or breakfasts or sides, as in questions 4, 5, 8) are captured in the recipe classification stub, at least in a first, simple fashion. For gluten-free and low-calorie, we carry basic inforamtion in the nutritional information, but do not yet provide corresponding categorizations. These categorizations could be added -- the appropriate place would be that they would be part of a refinement of the nutritional information module. The same holds for the notion of low-carb in the second question. Pot roast, as in question 2, and Chili as in question 3 are names of foods which are prepared following a recipe, i.e. it is the recipe name which holds this information. Under 100 calories is captured in nutritional information, though incompletely so, as the nutritional content of the final dish may need to be calculated from the ingredients and serving sizes. Currently, our ontology can list this only if the web page from which the recipe originates carries this information. 
The fact that a breakfast may be ``sweet'' cannot be captured currently. How to model this would need some contemplation -- in the end it is a subjective assessment, in a similar way in which ``low-calorie'' or ``simple'' would be a subjective assessment. On the other hand, given the use cases and the fact that recipes are retrieved from Web resources, this may be a case for simply adding some keyword tags obtained from the source.

\section{Creating OWL Files}\label{sec:rec-owl}

\emph{Step 5: Create OWL files.} 

Modeling up to this stage is usually done using paper, whiteboards, text documents. Only after we have created a solid modular model, we move to creating a data artefact in form of an OWL file which captures our ontology. 

One of the problems with using OWL, however, is that it does not natively support modularization in the sense in which we are presenting it. In order to preserve the modularization, one option is to make use of different namespaces for the different related modules, and if a class can be understood as belonging to two different modules, then we recommend to duplicate this class under different namespaces and to set these classes to be equivalent using an \texttt{owl:equivalentClass} axiom. A cleaner solution, rather than indicating modules using namespaces, is to make use of the Ontology Design Pattern Representation Language OPLa \cite{HitzlerGJKP17,ShimizuHH18} which is expressed fully in OWL -- however we do not go into further detail on this here.

We may choose to include mappings to external entities, e.g., alignments to other ontologies or to external ontology design patterns which were used as templates during the creation of our modules. Indeed, we recommend to keep such external models entirely separate from our own modules, by exclusively using local own, controlled namespaces within the modules, even if pieces from other ontologies are used verbatim. Instead, mappings to such external ontologies should be provided, and they should be provided as separate OWL files. The simple reason for this is that, once merged, it is hard to disentangle internal and external terms. Furthermore, external ontologies may change over time, and their axiomatizations or perspectives may not fit our model completely. By keeping the mappings separate, one can much more easily choose to opt into these mappings, or consult them only if needed.

The completed ontology should, of course, also be documented carfully. Documentation should reflect the modular structure.



\bigskip

Inspiration and some content for this tutorial was taken from \cite{SamKWGH14}. A similar basic introductory example can be found in \cite{chess-odpbook}. A report on a modular application ontology was published in \cite{KrisnadhiHJHACC15}.

\bibliographystyle{plain}
\bibliography{bib}

\end{document}